\documentclass[10pt,twocolumn,letterpaper]{article}
\usepackage[algorithms]{wacv}      
\usepackage{graphicx}
\usepackage{amsmath}
\usepackage{amssymb}
\usepackage{booktabs}
\usepackage[pagebackref,breaklinks,colorlinks]{hyperref}

\usepackage[capitalize]{cleveref}
\crefname{section}{Sec.}{Secs.}
\Crefname{section}{Section}{Sections}
\Crefname{table}{Table}{Tables}
\crefname{table}{Tab.}{Tabs.}

\def\wacvPaperID{}
\def\confName{WACV}
\def\confYear{2025}

\usepackage{comment}
\usepackage{siunitx}
\usepackage{caption}
\usepackage{subcaption}
\usepackage{geometry}
\usepackage[section]{placeins}
\usepackage{tcolorbox}
\usepackage{array}
\usepackage{adjustbox}
\usepackage{pdflscape}
\usepackage{float}
\usepackage{colortbl}

\begin{document}

\title{OPCap:Object-aware Prompting Captioning}

\author{
  Feiyang Huang\\
  South China Normal University\\
  {\tt\small flynn@m.scnu.edu.cn}
}
\maketitle

\begin{abstract}
	In the field of image captioning, the phenomenon where missing or nonexistent objects are used to explain an image is referred to as object bias (or hallucination). To mitigate this issue, we propose a target-aware prompting strategy. This method first extracts object labels and their spatial information from the image using an object detector. Then, an attribute predictor further refines the semantic features of the objects. These refined features are subsequently integrated and fed into the decoder, enhancing the model's understanding of the image context. Experimental results on the COCO and nocaps datasets demonstrate that OPCap effectively mitigates hallucination and significantly improves the quality of generated captions.
\end{abstract}

\section{Introduction}
Image captioning has attracted considerable attention in recent years as a cross-modal task. The goal of image captioning is to automatically generate descriptive text for a given image. This requires the model not only to understand the visual content within the image but also to convert this visual information into fluent and accurate language. Image captioning has broad application potential in areas such as assistive technology for the visually impaired, image search engine optimization, smart album management, and social media content analysis.

\begin{figure}
    \centering
    \caption{Examples of hallucination}
    \begin{minipage}[t]{0.45\textwidth}
        \begin{minipage}[c]{0.45\textwidth}
            \includegraphics[width=\linewidth,height=\linewidth]{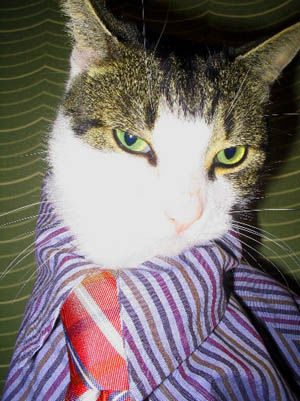}
        \end{minipage}%
        \hspace{0.05\textwidth}%
        \begin{minipage}[c]{0.5\textwidth}
            \begin{tcolorbox}[
                    colback=blue!15,
                    coltext=black,
                    width=\linewidth,
                    bottom=0.5mm,top=0.5mm,
                    left=1mm,right=1mm,
                    boxsep=0pt,
                    colframe=white
                ]
                \small\textbf{objects:} cat, tie
            \end{tcolorbox}
            \vspace{2mm}
            \begin{tcolorbox}[
                    colback=red!15,
                    coltext=black,
                    width=\linewidth,
                    bottom=0.5mm,top=0.5mm,
                    left=1mm,right=1mm,
                    boxsep=0pt,
                    colframe=white
                ]
                \small \textbf{With hallucination:}  a close up of a \textcolor{red}{\textbf{person}} wearing a tie.
            \end{tcolorbox}
            \vspace{2mm}
            \begin{tcolorbox}[
                    colback=green!15,
                    coltext=black,
                    width=\linewidth,
                    bottom=0.5mm,top=0.5mm,
                    left=1mm,right=1mm,
                    boxsep=0pt,
                    colframe=white
                ]
                \small \textbf{Without hallucination:}  a close up of a cat wearing a tie
            \end{tcolorbox}
        \end{minipage}
    \end{minipage}

    \vspace{0.5cm}  

    \noindent\leaders\hbox{\rule{.4em}{.4pt}\hspace{.4em}}\hfill\mbox{}\par

    \vspace{0.5cm}  

    \begin{minipage}[t]{0.45\textwidth}
        \begin{minipage}[c]{0.45\textwidth}
            \includegraphics[width=\linewidth,height=\linewidth]{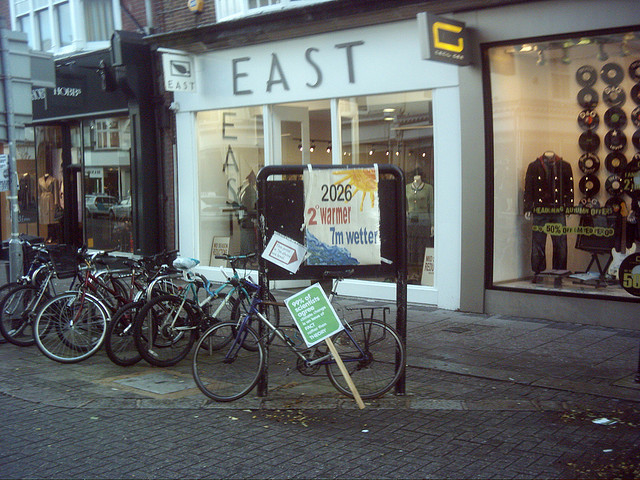}
        \end{minipage}%
        \hspace{0.05\textwidth}%
        \begin{minipage}[c]{0.5\textwidth}
            \begin{tcolorbox}[
                    colback=blue!15,
                    coltext=black,
                    width=\linewidth,
                    bottom=0.5mm,top=0.5mm,
                    left=1mm,right=1mm,
                    boxsep=0pt,
                    colframe=white
                ]
                \small \textbf{objects:} bicycle
            \end{tcolorbox}
            \vspace{2mm}
            \begin{tcolorbox}[
                    colback=red!15,
                    width=\linewidth,
                    coltext=black,
                    bottom=0.5mm,top=0.5mm,
                    left=1mm,right=1mm,
                    boxsep=0pt,
                    colframe=white
                ]
                \small \textbf{With hallucination:} a group of \textcolor{red}{\textbf{people}}  sitting on a street.
            \end{tcolorbox}
            \vspace{2mm}
            \begin{tcolorbox}[
                    colback=green!15,
                    width=\linewidth,
                    coltext=black,
                    bottom=0.5mm,top=0.5mm,
                    left=1mm,right=1mm,
                    boxsep=0pt,
                    colframe=white
                ]
                \small \textbf{Without hallucination:} a bicycle parked on a bench in front of a building.
            \end{tcolorbox}
        \end{minipage}
    \end{minipage}
    \label{fig:Examples of hallucination}
\end{figure}

Despite significant progress in image captioning, several challenges remain, with one of the most prominent being object hallucination \cite{rohrbach2019objecthallucinationimagecaptioning}. As shown in Figure \ref{fig:Examples of hallucination}, objects irrelevant to the image appear in the description. Such a phenomenon not only affects the accuracy of the captions but also risks conveying misleading information. Most existing methods rely on deep learning-based encoder-decoder architectures, which map image features to textual descriptions through end-to-end training \cite{Anderson_2018_CVPR, Qin_2019_CVPR, Huang_2019_ICCV, mokady2021clipcapclipprefiximage}. However, these approaches often fail to fully consider the presence or absence of specific objects in the image, making them prone to object hallucination.

In addition, many advanced image captioning methods rely on large-scale image-text datasets for incremental pretraining or use additional pre-trained language models to initialize and fine-tune the caption generation decoder. While these approaches can improve model performance, they also incur significant resource costs, including computational resources and time. For resource-constrained environments or applications, such methods may be impractical, limiting their potential for widespread use.

In light of the challenges mentioned above, our primary goal is to explore a simple and effective method for generating accurate and high-quality image captions without relying on additional pre-trained language models or large-scale datasets. To achieve this, we propose a target-aware prompting strategy (Object-aware Prompting Captioning, OPCap). This method integrates an object detector and an attribute predictor to enhance the model's understanding of the image content. 

Specifically, OPCap uses a pre-trained object detector to identify key object labels and their spatial information from the image. It then extracts the corresponding image regions based on the spatial information and feeds them into the attribute predictor to obtain the object's attributes. These object labels and attributes are combined with the features extracted by the image encoder and passed to the decoder. This approach enables the model to better focus on the actual objects in the image during caption generation, effectively reducing the occurrence of object bias.

We conducted experiments and ablation studies on two widely used datasets, COCO and nocaps. Our main contributions are as follows:
\begin{itemize}
	\item We propose an object-aware prompting method that can be applied to any image captioning model.
	\item Our method is resource-efficient and suitable for resource-constrained scenarios, without requiring additional pre-trained models or large-scale datasets.
	\item Experimental results show that OPCap significantly reduces object hallucination and improves the overall quality of generated captions.
\end{itemize}

\section{Related Work}
\subsection{Image captioning models}
Image captioning models are typically composed of an image encoder and a text decoder, corresponding to the visual and textual modalities, respectively. The image encoder extracts visual features, while the text decoder generates captions based on these features. Researchers have continually improved these components. Early work \cite{conf/cvpr/VinyalsTBE15} employed convolutional neural network (CNN) as image encoders and long short-term memory network (LSTM \cite{10.1162/neco.1997.9.8.1735}) as text decoders. However, LSTMs often struggle with capturing long-range dependencies due to issues like vanishing or exploding gradients, limiting their ability to retain long-term historical information. To address this, attention mechanisms were introduced into LSTMs \cite{pmlr-v37-xuc15}, enabling the model to focus on different regions of an image and produce more accurate captions. 

With the success of Transformer (\cite{10.5555/3295222.3295349}) models in natural language processing (NLP), many image captioning models have adopted Transformer-based architectures \cite{Xian2022Adaptive, Cao2022Vision-Enhanced, Shen2020Remote}. These models leverage attention mechanisms to process image and text data simultaneously, significantly improving the performance of caption generation.

\subsection{Hallucnation in models}
Hallucination issues often stem from insufficient understanding of image content by the model or biases in the training data. For instance, the model may overly rely on language priors, generating descriptions of common objects that are irrelevant to the actual image. In Figure 1, an image of a cat wearing a tie is mistakenly described as a person wearing a tie. This occurs because in the training data, the combination of "person" and "tie" is more frequent than "cat" and "tie," leading the model to generate descriptions related to people. Furthermore, the lack of fine-grained alignment between visual and textual representations exacerbates hallucination, resulting in captions that are semantically plausible but actually incorrect in terms of the image content.

Research by Anna Rohrbach et al. \cite{rohrbach2018object} suggests that models achieving higher scores on image caption evaluation metrics do not necessarily exhibit lower hallucination rates. For instance, Steven et al. \cite{Gao_2019_CVPR} trained model with self-critical loss, which improve these metrics, may inadvertently lead to an increase in hallucinations. Furthermore, they note that current evaluation metrics primarily measure the similarity between generated captions and ground truth, but fail to account for the relevance of the captions to the actual image. To address this limitation, they introduce two new metrics, CHAIRs and CHAIRi, which evaluate the degree of object mislabeling in captions.

\subsection{Hallucination reduction methods}
Ali Furkan Biten et al. \cite{Biten2021LetTB} proposed a method that incorporates object labels as input along with data augmentation. In this approach, an object detector identifies objects in the image, and the detected object labels, after embedding, are concatenated with the features extracted by an image encoder. This combined input is then used for further processing. The data augmentation is achieved by utilizing an object co-occurrence matrix to aid decision-making, replacing objects in the ground truth, reflecting a regularization strategy. 

Lei WANG et al. \cite{wang2024mitigating} introduced the ReCaption framework, which leverages ChatGPT to rewrite captions and fine-tunes LVLMs to mitigate fine-grained hallucinations. The framework operates in two stages. The first stage focuses on keyword extraction, aiming to retain essential descriptive information that is closely tied to the specific image. The second stage generates new captions based on the extracted keywords from the first stage. This method improves the implicit fine-grained alignment between the input image and the generated caption, which helps reduce fine-grained object hallucinations and enhances the overall quality of the caption.

Compared to these methods, the proposed OPCap does not rely on external language models like ChatGPT, thereby avoiding potential noise and computational overhead. Additionally, OPCap combines object detection with attribute prediction to explicitly model the relationships between objects and their attributes, providing a more fine-grained understanding of the image content. This is crucial for reducing both coarse-grained and fine-grained hallucinations. Moreover, both the object detector and attribute extractor are trained or fine-tuned on the COCO dataset, making efficient use of the dataset's resources.

\section{Method}
\begin{figure*}
	\centering
	\includegraphics[width=0.85\textwidth]{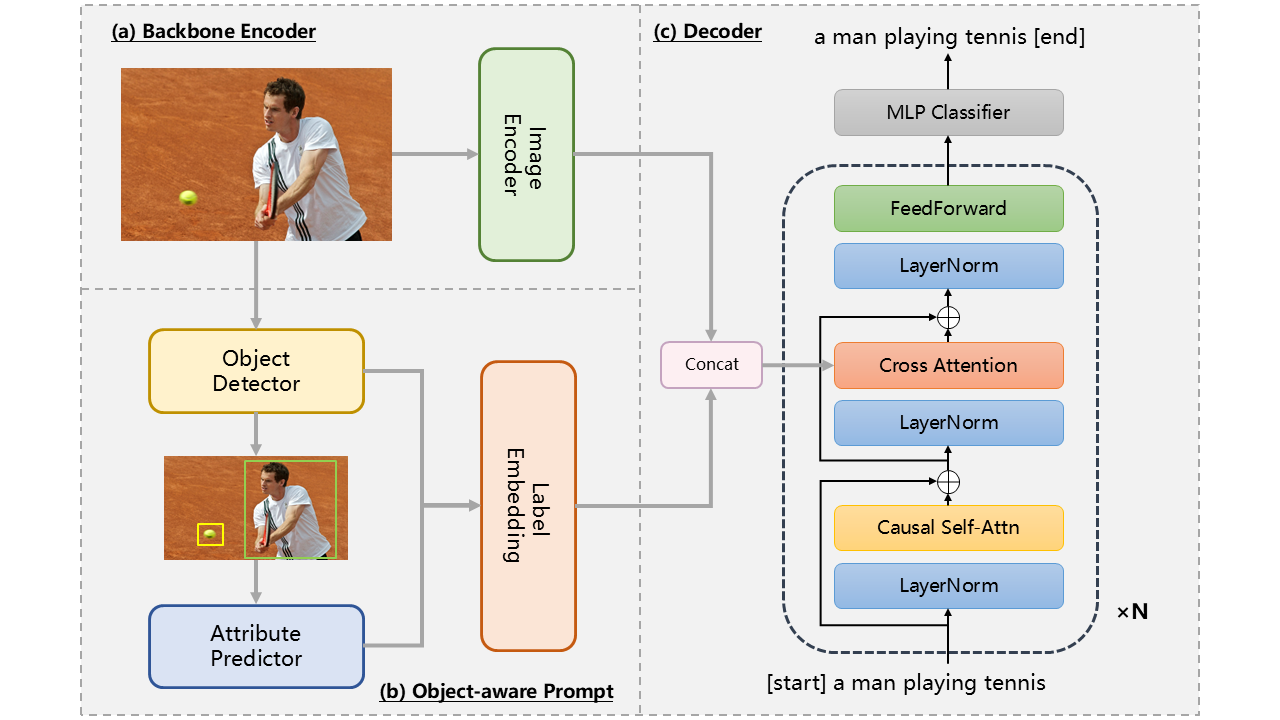}
	\caption{OPCap Architecture: The architecture consists of three modules, including the image encoder, object detector + attribute predictor, and text decoder.}
	\label{fig:OPCap Architecture}
\end{figure*}
We propose a new strategy called OPCap, which aims to reduce hallucinations in image captioning by explicitly incorporating object-level information. As shown in Figure \ref{fig:OPCap Architecture}, the method mainly includes four steps: image encoding, object detection, attribute prediction and decoding. By integrating detected objects and their attributes into the caption generation process, OPCap enhances the model's understanding of the image context without relying on external language models.

\subsection{Object-aware Prompting}
A common issue in datasets is the imbalance in the number of images across different object categories. For instance, in the Flickr30K dataset, samples containing people are approximately 10 times more frequent than those without people. This imbalance may cause the model to overfit to frequent categories while struggling to recognize and describe rare categories effectively. To address this issue, an object detector can be introduced as an auxiliary tool to provide more accurate object information.

Object detection is a classic task in computer vision, and leveraging pretrained models can significantly improve efficiency. To mitigate the computational burden of the object-aware prompting module, we employ yolos-tiny \cite{DBLP:journals/corr/abs-2106-00666} as the object detection model. After passing the image through the object detector, we obtain the labels and positions of the objects within the image. These object regions are then cropped and used as input for the attribute predictor.

To further refine the object information, we introduce an attribute predictor, which takes the cropped image regions as input and predicts detailed attributes such as color, shape, and state.

Attribute prediction is a multi-label classification task. For this, we built a classification model based on the VAW \cite{Pham_2021_CVPR} dataset. Specifically, we employed CLIP \cite{pmlr-v139-radford21a} as the image encoder and integrated it with a multilayer perceptron (MLP) classifier.

During prediction, we select the top-k attributes after applying sigmoid activation. The generated labels are then concatenated into a custom token sequence. For example, in the image shown in Figure \ref{fig:OPCap Architecture}, with $k=2$, the corresponding token sequence is: \\
\texttt{[OBJ] person [ATTR] gray [ATTR] swinging} \\
\texttt{[OBJ] sports ball [ATTR] small [ATTR] rounded}

After embedding, the token sequence is combined with the image features and passed as input to the decoder.

\subsection{Caption Generation and Training}

In the model, the image is separately fed into the image encoder and the object detector. The image encoder outputs $m$ feature vectors, while the object detector generates a list containing $o$ objects. The attribute predictor then predicts $k$ attributes for each object. To differentiate these pieces of information, we use custom special tokens to separate the object-related information. Finally, through the embedding layer, this information is mapped into $2 \times (o + k)$ feature vectors.
\begin{equation}
	x^{i}_{1},...,x^{i}_{m} = \text{ImageEncoder}(I^{i})
\end{equation}
\begin{equation}
	objs^i, boxes^i = \text{ObjectDetector}(\text{I}^i)
\end{equation}
\begin{equation}
	attrs^i = \text{AttributePredictor}(objs^i, boxes^i)
\end{equation}
\begin{equation}
	e^{i}_{1},...,e^{i}_{2 \times(o+k)} = \text{Embedding}(objs^i, attrs^i)
\end{equation}
Here, $I^{i}$ represents the $i$-th sample, while $objs^i$ and $boxes^i$ refer to the object labels and their corresponding bounding boxes, respectively. $attrs^i$ represents the attributes associated with each object. Both $x^{i}$ and $e^{i}$ are mapped to the same feature dimension, $d_{model}$, ensuring they share the same embedding space. Subsequently, these two features are concatenated as:
\begin{equation}
	Z^{i} = x^{i}_{1},..,x^{i}_{m},e^{i}_{1},...,e^{i}_{2 \times(o+k)}
\end{equation}

The caption generation process is autoregressive. Since LSTM has difficulty handling the joint features of both image and object information, we use a Transformer in the decoder. To prevent the model from overly relying on linguistic priors and to improve its robustness, we first apply random dropout to the tokens in the caption before feeding the embedded caption into the causal attention layer. Let $C^i$ denote the processed input caption, $Z^i$ be the key and value for the Transformer, and $C^i$ be the query. The output of the Transformer is then passed through an MLP classifier to complete the decoding process:
\begin{equation}
	\text{Output} = \text{MLP}(\text{Transformer}(C^i, Z^i))
\end{equation}

We use LM as the training loss and the objective function is:
\begin{equation}
	\mathcal{L}_{\text{LM}}(\theta) = -\sum_{t=1}^{T} \log P(w_i | w_{<t}, I; \theta)
\end{equation}
where \(\theta\) denotes the model parameters, \(T\) is the length of the caption, \(w_t\) is the \(t\)-th word in the caption, \(w_{<t}\) represents all words before the \(t\)-th word, \(I\) is the input image, and \(P(w_t | w_{<t}, I; \theta)\) is the probability of generating the word \(w_t\) given the image \(I\), the preceding words \(w_{<t}\), and the model parameters \(\theta\).

\section{Experiments}
\subsection{Dataset, Evaluation Metrics and Baselines}
\textbf{Dataset:} We use the widely adopted image caption dataset MSCOCO \cite{lin2015microsoftcococommonobjects}, utilizing the latest 2017 version available on the official website. The dataset consists of 118,287 images, with 118,287 images designated for training and 5,000 images set aside for validation and testing.

\textbf{Evualation Metrics:} To evaluate the quality of the generated image captions, we use standard automatic evaluation metrics, including CIDEr \cite{vedantam2015cider}, BLEU \cite{papineni2002bleu}, METEOR \cite{banerjee2005meteor}, and SPICE \cite{anderson2016spice}. Additionally, to assess improvements in mitigating hallucination, we employ the hallucination metrics CHAIRs and CHAIRi \cite{rohrbach2018object}, which evaluate hallucinations at the sentence and object levels, respectively.

\textbf{Baselines:} Most image captioning models adopt an Encoder-Decoder architecture, with key differences between models lying in the design of the image encoder, text decoder, and multimodal fusion module. In our experiments, we selected the following models based on different technical approaches for comparison:

\begin{itemize}
    \item \textbf{UpDown model \cite{anderson2018bottom}}: Generates captions using the Bottom-Up and Top-Down attention mechanisms, with Faster R-CNN extracting salient region features and LSTM generating the captions.
    \item \textbf{AoA model \cite{huang2019attention}}: Introduces a dual-attention mechanism, adding a second attention module on top of the first to refine the attention weights and capture the relationship between image and text more precisely.
    \item \textbf{ViTGPT2}: Uses ViT to extract global image features and leverages the powerful generative capabilities of GPT-2 to produce high-quality captions.
    \item \textbf{CLIP-based models \cite{radford2021learningtransferablevisualmodels}}: Combine Transformer as a decoder, enabling better alignment between the image and text semantic spaces, resulting in more accurate captions.
\end{itemize}

\subsection{Qualitative Analysis}
\begin{table*}
    \centering
    \renewcommand{\arraystretch}{1.1}
    \caption{Comparison of different model outputs. The "+OP" notation indicates models that incorporate our proposed object-aware prompting method.}
    \vspace{0.2cm}
    \begin{adjustbox}{width=\textwidth}
        \begin{tabular}{l>{\raggedright\arraybackslash}p{3cm}>{\raggedright\arraybackslash}p{3cm}>{\raggedright\arraybackslash}p{3cm}>{\raggedright\arraybackslash}p{3cm}>{\raggedright\arraybackslash}p{3cm}}
            & \includegraphics[width=3cm]{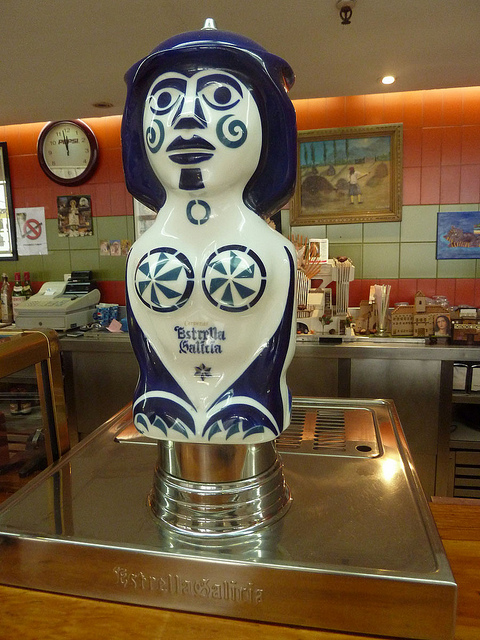}
            & \includegraphics[width=3cm]{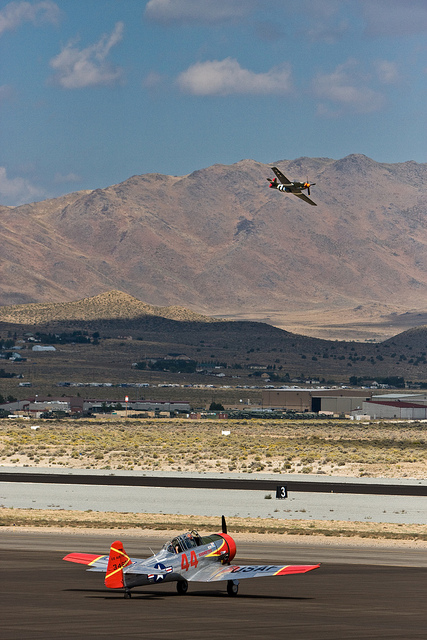}
            & \includegraphics[width=3cm]{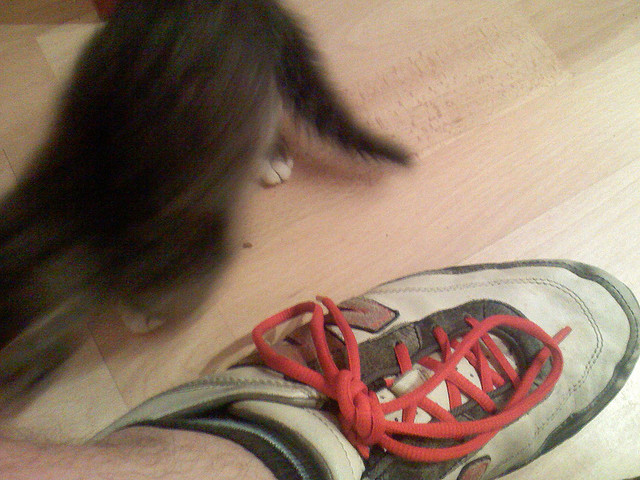}
            & \includegraphics[width=3cm]{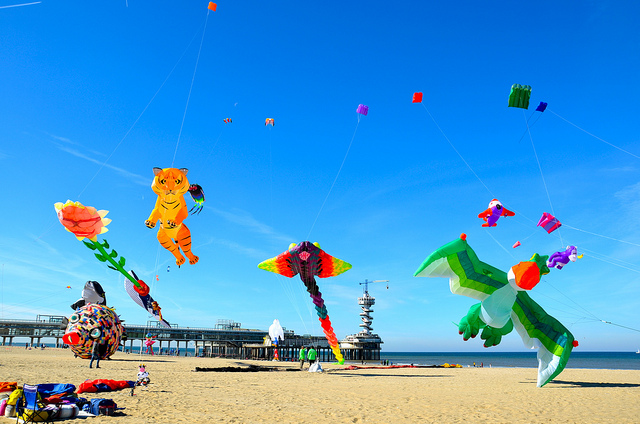}
            & \includegraphics[width=3cm]{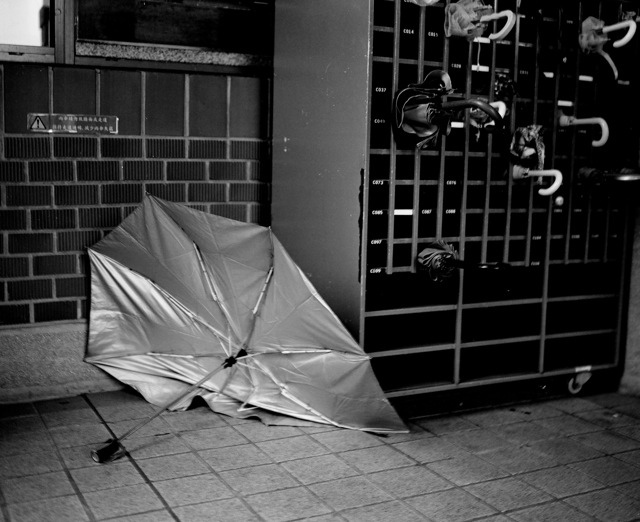} \\
            \footnotesize
            EffNet &  a close up of a \textcolor{red}{person} in a room. & a group of \textcolor{red}{people} on top of a beach. & a bedroom with a bed and a bed. & a group of people flying kites on a beach. & there is a picture of a man in the air \\ 
            \hline
            EffNet+OP &  a close up of a vase on a table & a plane is flying over a sandy beach. & a cat laying on a floor next to a pair of shoes. & a group of people flying kites in the sky. & a black and white photo of a bed in a room. \\ 
            \hline
            CLIP-base &  a white and blue vase sitting on a counter. & a \textcolor{red}{man} flying through the air while riding a motorcycle. & a close up of a \textcolor{red}{person} playing with shoes & a group of people flying kites on a beach. & a black and white photo of a \textcolor{red}{person} holding an umbrella. \\ 
            \hline
            CLIP-base+OP &  a statue of a woman in a kitchen. & a large air plane flying in the air on a mountain. & a black and white cat playing with a pair of shoes. & a bunch of kites flying over a beach. & a black and white photo of an umbrella on a bench. \\ 
            \hline
            CLIP-large &  a statue of a woman in a restaurant. & a plane flying in the sky over a field. & a cat standing next to a pair of sneakers. & a group of people on a beach flying kites. & a black and white photo of an umbrella on a sidewalk. \\ 
            \hline
            CLIP-large+OP &  a statue of a statue on a counter. & a plane flying over an airplane in a desert. & a cat looking at a pair of sneakers. & a group of people standing on top of a sandy beach. & an umbrella sitting on top of a floor. \\ 
            \hline
        \end{tabular}
    \end{adjustbox}
    \label{tab:comparison}
\end{table*}

As shown in Table \ref{tab:comparison}, we randomly selected 5 samples from the COCO dataset to compare the performance of different models. The results demonstrate that models using our proposed OPCap method generate image captions with improved detail capture and semantic accuracy, while also alleviating hallucination issues. 

However, we also identified some limitations. Specifically, the effectiveness of the target-aware prompt strategy is partially dependent on the performance of the object detector. For instance, in the fourth image, the object detector failed to detect the crowd, resulting in omissions in the generated caption. 

Moreover, during training, we froze the parameters of the image encoder entirely, and different image encoders significantly affect the generated captions. Overall, this variation is positively correlated with the complexity of the image encoder.

\subsection{Quantitative analysis}
\begin{table*}
	\centering
	\caption{The evaluation results of different models on traditional image caption metrics for the COCO and NoCaps datasets. The "+OP" notation indicates models that incorporate our proposed object-aware prompting method.}
	\begin{subtable}{\textwidth}
		\centering
		\caption{Evaluation results on nocaps. R : ROUGE-L metric, C : CIDEr, S : SPICE.}
        \resizebox{\textwidth}{!}{
		\begin{tabular*}{\linewidth}{@{\extracolsep{\fill}}l|ccc|ccc|ccc}
			\hline
			& \multicolumn{3}{c}{In-domain} & \multicolumn{3}{c}{Near-domain} & \multicolumn{3}{c}{Out-of-domain} \\
			Model & R $\uparrow$ & C $\uparrow$ & S $\uparrow$ & R & C & S & R & C & S \\
			\hline
			\hline
            EffNet & 41.89 & 29.25 & 5.5 & 40.11 & 20.94 & 4.37 & 37.37 & 9.91 & 2.66 \\
            EffNet+OP & 43.02 & 37.62 & 7.09 & 41.8 & 29.56 & 5.81 & 37.73 & 13.53 & 3.47 \\
            ViTGPT2 & 51.51 & 65.92 & 10.67 & 51.63 & 63.38 & 10.48 & 45.43 & 43.1 & 8.17 \\
            CLIP-base & 54.13 & 71.97 & 11.07 & 52.37 & 64.34 & 10.07 & 44.71 & 44.33 & 7.39 \\
            CLIP-base+OP & 53.28 & 72.05 & 11.14 & 51.69 & 64.28 & 10.18 & 44.99 & 44.47 & 7.74 \\
            CLIP-large & \textbf{ 54.46 } & 74.65 & \textbf{ 11.59 } & \textbf{ 53.35 } & \textbf{ 71.39 } & \textbf{ 10.76 } & \textbf{ 46.7 } & \textbf{ 57.14 } & \textbf{ 8.67 } \\
            CLIP-large+OP & 53.68 & \textbf{ 75.56 } & 11.27 & 52.97 & 70.57 & 10.68 & 44.2 & 47.02 & 8.21 \\
			\hline
		\end{tabular*}
        }
		\label{tab:Evaluation results on nocaps}
	\end{subtable}
	\vspace{1em}

	\begin{subtable}{\textwidth}
		\centering
		\renewcommand{\arraystretch}{1.1}
		\caption{Evaluation results on COCO}
        \resizebox{\textwidth}{!}{
            \begin{tabular}{l|c|c|c|c|c|c|c}
                \hline
                Model & Bleu-4 $\uparrow$  & METEOR $\uparrow$ & ROUGE-L $\uparrow$ & CIDEr $\uparrow$ & SPICE $\uparrow$ & CHAIRs $\downarrow$ & CHAIRi $\downarrow$ \\ 
                \hline
                \hline
                EffNet & 20.73 & 18.13 & 44.74 & 55.95 & 11.21  & \cellcolor{gray!20} 40.84 & \cellcolor{gray!20} 33.04 \\
                EffNet+OP & 22.64 & 20.76 & 46.72 & 68.09 & 13.65  & \cellcolor{gray!20} 32.78 & \cellcolor{gray!20} 26.04 \\
                ViTGPT2 & 33.23 & 27.09 & 55.15 & 108.61 & 20.03  & \cellcolor{gray!20} 11.04 & \cellcolor{gray!20} 7.63 \\
                AoA & 33.7 & - & 27.4 & 111.0 & 20.6 & \cellcolor{gray!20}9.1 & \cellcolor{gray!20}6.2 \\
                UpDown & 33.2 & - & 26.9 & 108.4 & 20.0 & \cellcolor{gray!20}10.1 & \cellcolor{gray!20}6.9 \\
                CLIP-base & 34.12 & 26.72 & 55.27 & 107.65 & 19.65  & \cellcolor{gray!20} 11.1 & \cellcolor{gray!20} 8.21 \\
                CLIP-base+OP & 34.04 & 26.74 & 55.19 & 107.72 & 19.57  & \cellcolor{gray!20} 10.16 & \cellcolor{gray!20} 7.33 \\
                CLIP-large & \textbf{ 36.64 } & \textbf{ 28.21 } & \textbf{ 56.93 } & \textbf{ 117.39 } & \textbf{ 21.12 }  & \cellcolor{gray!20} 6.34 & \cellcolor{gray!20} 4.47 \\
                CLIP-large+OP & 35.96 & 27.47 & 56.52 & 115.59 & 20.8  & \cellcolor{gray!20} \textbf{5.8} & \cellcolor{gray!20} \textbf{4.27} \\
            \end{tabular}
        }
		\label{tab:Evaluation results on COCO}
	\end{subtable}
\end{table*}

\textbf{Traditional Image Caption Metrics:} We conducted evaluations on the nocaps \cite{agrawal2019nocaps} and COCO datasets. The nocaps dataset is designed to assess the generalization ability of image caption models to unseen categories and concepts. It consists of three main subsets: in-domain, near-domain, and out-of-domain, which progressively test the model's ability to describe images involving new categories. Additionally, we performed evaluations on the COCO dataset, which is a widely used image caption dataset containing images from a variety of categories. It is commonly employed to train and assess the diversity and accuracy of image caption models.

As shown in Table \ref{tab:Evaluation results on nocaps} and \ref{tab:Evaluation results on COCO}, when using a relatively weak pre-trained image encoder, models with target-aware prompts generally exhibit improvements across traditional evaluation metrics. In contrast, models based on CLIP show the opposite trend. This could be because the image encoder in CLIP already captures relevant information about the target and its attributes in the extracted features. On the hallucination evaluation metrics, CHAIRi and CHAIRs, models using target-aware prompts also show varying degrees of improvement. This aligns with the findings in [], suggesting that models performing well on traditional metrics do not necessarily achieve the same results in hallucination evaluations.

\begin{table}[h]
    \centering
    \renewcommand{\arraystretch}{1.1}
    \caption{CLIP Vote Results}
    \begin{tabular}{lcc}
        \hline
        Model & Vote $\uparrow$ & Param (M)\\
        \hline
        \hline
        CLIP-base+OP & 1391 & \textbf{56.51} \\
        CLIP-base & 1218 & \textbf{56.51}\\
        CLIP-large+OP & 1093 & 103.63\\
        CLIP-large & 1250 & 103.63 \\
        EffNet+OP & 330 & 235.6 \\
        EffNet & 163 & 235.6 \\
        ViTGPT2 & 739 & 239.20\\
        \hline
    \end{tabular}
    \label{tab:Clip Vote}
\end{table}

\textbf{CLIP-based Metric:} CLIP, proposed by OpenAI, is a multimodal model trained on a large number of image-text pairs using contrastive learning. It maps images and texts into a shared embedding space, demonstrating strong cross-modal understanding capabilities. In \cite{Rotstein_2024_WACV}, the authors evaluate which of two descriptions (original and generated) is preferred for a given image. Inspired by this approach, we used CLIP to compute the similarity between an image and descriptions generated by different models. The model with the highest similarity score for a given image receives one point (if multiple descriptions have the same similarity score, they all receive points), resembling a voting process, which we term as CLIP Vote.

Traditional metrics typically focus on the similarity between the generated caption and the reference caption. However, they somewhat overlook whether the caption accurately and comprehensively captures the content of the image. On the other hand, CLIP Vote, as a no-reference metric, helps address this issue to some extent. As shown in Table \ref{tab:Clip Vote}, we randomly selected 5000 images from the COCO dataset for testing. Surprisingly, the base version of CLIP received the highest number of votes, while models incorporating the OPCap method consistently received more votes.

\section{Conclusion}
We observed that image caption models tend to produce inaccurate descriptions or hallucinations when handling uncommon images. We hypothesize that this issue is not solely due to the limitations of the dataset. In response, we improved the model architecture by proposing an embedding method that combines an object detector and an attribute predictor. This approach fuses image features with object information, providing more effective prompt signals to the text decoder.

To evaluate the effectiveness of the OPCap method, we conducted several experiments on the nocaps and COCO datasets. The results demonstrate that the target-aware strategy and the fusion module significantly enhance model performance, validating the improvements introduced by the OPCap method.

{\small
  \bibliographystyle{unsrt}  
	\bibliography{01_introduction_ref, 02_related_works_ref, 03_method_ref, 04_experiments_ref}

\begin{thebibliography}{10}

\bibitem{rohrbach2019objecthallucinationimagecaptioning}
Anna Rohrbach, Lisa~Anne Hendricks, Kaylee Burns, Trevor Darrell, and Kate Saenko.
\newblock Object hallucination in image captioning, 2019.

\bibitem{Anderson_2018_CVPR}
Peter Anderson, Xiaodong He, Chris Buehler, Damien Teney, Mark Johnson, Stephen Gould, and Lei Zhang.
\newblock Bottom-up and top-down attention for image captioning and visual question answering.
\newblock In {\em Proceedings of the IEEE Conference on Computer Vision and Pattern Recognition (CVPR)}, June 2018.

\bibitem{Qin_2019_CVPR}
Yu~Qin, Jiajun Du, Yonghua Zhang, and Hongtao Lu.
\newblock Look back and predict forward in image captioning.
\newblock In {\em Proceedings of the IEEE/CVF Conference on Computer Vision and Pattern Recognition (CVPR)}, June 2019.

\bibitem{Huang_2019_ICCV}
Lun Huang, Wenmin Wang, Jie Chen, and Xiao-Yong Wei.
\newblock Attention on attention for image captioning.
\newblock In {\em Proceedings of the IEEE/CVF International Conference on Computer Vision (ICCV)}, October 2019.

\bibitem{mokady2021clipcapclipprefiximage}
Ron Mokady, Amir Hertz, and Amit~H. Bermano.
\newblock Clipcap: Clip prefix for image captioning, 2021.

\bibitem{conf/cvpr/VinyalsTBE15}
Oriol Vinyals, Alexander Toshev, Samy Bengio, and Dumitru Erhan.
\newblock Show and tell: A neural image caption generator.
\newblock In {\em CVPR}, pages 3156--3164. IEEE Computer Society, 2015.

\bibitem{10.1162/neco.1997.9.8.1735}
Sepp Hochreiter and J\"{u}rgen Schmidhuber.
\newblock Long short-term memory.
\newblock {\em Neural Comput.}, 9(8):1735–1780, November 1997.

\bibitem{pmlr-v37-xuc15}
Kelvin Xu, Jimmy Ba, Ryan Kiros, Kyunghyun Cho, Aaron Courville, Ruslan Salakhudinov, Rich Zemel, and Yoshua Bengio.
\newblock Show, attend and tell: Neural image caption generation with visual attention.
\newblock In Francis Bach and David Blei, editors, {\em Proceedings of the 32nd International Conference on Machine Learning}, volume~37 of {\em Proceedings of Machine Learning Research}, pages 2048--2057, Lille, France, 07--09 Jul 2015. PMLR.

\bibitem{10.5555/3295222.3295349}
Ashish Vaswani, Noam Shazeer, Niki Parmar, Jakob Uszkoreit, Llion Jones, Aidan~N. Gomez, \L{}ukasz Kaiser, and Illia Polosukhin.
\newblock Attention is all you need.
\newblock In {\em Proceedings of the 31st International Conference on Neural Information Processing Systems}, NIPS'17, page 6000–6010, Red Hook, NY, USA, 2017. Curran Associates Inc.

\bibitem{Xian2022Adaptive}
Tiantao Xian, Zhixin Li, Zhenjun Tang, and Huifang Ma.
\newblock Adaptive path selection for dynamic image captioning.
\newblock {\em IEEE Transactions on Circuits and Systems for Video Technology}, 32:5762--5775, 2022.

\bibitem{Cao2022Vision-Enhanced}
Shan Cao, Gaoyun An, Zhenxing Zheng, and Zhiyong Wang.
\newblock Vision-enhanced and consensus-aware transformer for image captioning.
\newblock {\em IEEE Transactions on Circuits and Systems for Video Technology}, 32:7005--7018, 2022.

\bibitem{Shen2020Remote}
Xiangqing Shen, Bing Liu, Yong Zhou, and Jiaqi Zhao.
\newblock Remote sensing image caption generation via transformer and reinforcement learning.
\newblock {\em Multimedia Tools and Applications}, 79:26661 -- 26682, 2020.

\bibitem{rohrbach2018object}
Anna Rohrbach, Lisa~Anne Hendricks, Kaylee Burns, Trevor Darrell, and Kate Saenko.
\newblock Object hallucination in image captioning.
\newblock {\em arXiv preprint arXiv:1809.02156}, 2018.

\bibitem{Gao_2019_CVPR}
Junlong Gao, Shiqi Wang, Shanshe Wang, Siwei Ma, and Wen Gao.
\newblock Self-critical n-step training for image captioning.
\newblock In {\em Proceedings of the IEEE/CVF Conference on Computer Vision and Pattern Recognition (CVPR)}, June 2019.

\bibitem{Biten2021LetTB}
Ali~Furkan Biten, Llu{\'i}s~G{\'o}mez i~Bigorda, and Dimosthenis Karatzas.
\newblock Let there be a clock on the beach: Reducing object hallucination in image captioning.
\newblock {\em 2022 IEEE/CVF Winter Conference on Applications of Computer Vision (WACV)}, pages 2473--2482, 2021.

\bibitem{wang2024mitigating}
Lei Wang, Jiabang He, Shenshen Li, Ning Liu, and Ee-Peng Lim.
\newblock Mitigating fine-grained hallucination by fine-tuning large vision-language models with caption rewrites.
\newblock In {\em International Conference on Multimedia Modeling}, pages 32--45. Springer, 2024.

\bibitem{DBLP:journals/corr/abs-2106-00666}
Yuxin Fang, Bencheng Liao, Xinggang Wang, Jiemin Fang, Jiyang Qi, Rui Wu, Jianwei Niu, and Wenyu Liu.
\newblock You only look at one sequence: Rethinking transformer in vision through object detection.
\newblock {\em CoRR}, abs/2106.00666, 2021.

\bibitem{Pham_2021_CVPR}
Khoi Pham, Kushal Kafle, Zhe Lin, Zhihong Ding, Scott Cohen, Quan Tran, and Abhinav Shrivastava.
\newblock Learning to predict visual attributes in the wild.
\newblock In {\em Proceedings of the IEEE/CVF Conference on Computer Vision and Pattern Recognition (CVPR)}, pages 13018--13028, June 2021.

\bibitem{pmlr-v139-radford21a}
Alec Radford, Jong~Wook Kim, Chris Hallacy, Aditya Ramesh, Gabriel Goh, Sandhini Agarwal, Girish Sastry, Amanda Askell, Pamela Mishkin, Jack Clark, Gretchen Krueger, and Ilya Sutskever.
\newblock Learning transferable visual models from natural language supervision.
\newblock In Marina Meila and Tong Zhang, editors, {\em Proceedings of the 38th International Conference on Machine Learning}, volume 139 of {\em Proceedings of Machine Learning Research}, pages 8748--8763. PMLR, 18--24 Jul 2021.

\bibitem{lin2015microsoftcococommonobjects}
Tsung-Yi Lin, Michael Maire, Serge Belongie, Lubomir Bourdev, Ross Girshick, James Hays, Pietro Perona, Deva Ramanan, C.~Lawrence Zitnick, and Piotr Dollár.
\newblock Microsoft coco: Common objects in context, 2015.

\bibitem{vedantam2015cider}
Ramakrishna Vedantam, C~Lawrence~Zitnick, and Devi Parikh.
\newblock Cider: Consensus-based image description evaluation.
\newblock In {\em Proceedings of the IEEE conference on computer vision and pattern recognition}, pages 4566--4575, 2015.

\bibitem{papineni2002bleu}
Kishore Papineni, Salim Roukos, Todd Ward, and Wei-Jing Zhu.
\newblock Bleu: a method for automatic evaluation of machine translation.
\newblock In {\em Proceedings of the 40th annual meeting of the Association for Computational Linguistics}, pages 311--318, 2002.

\bibitem{banerjee2005meteor}
Satanjeev Banerjee and Alon Lavie.
\newblock Meteor: An automatic metric for mt evaluation with improved correlation with human judgments.
\newblock In {\em Proceedings of the acl workshop on intrinsic and extrinsic evaluation measures for machine translation and/or summarization}, pages 65--72, 2005.

\bibitem{anderson2016spice}
Peter Anderson, Basura Fernando, Mark Johnson, and Stephen Gould.
\newblock Spice: Semantic propositional image caption evaluation.
\newblock In {\em Computer Vision--ECCV 2016: 14th European Conference, Amsterdam, The Netherlands, October 11-14, 2016, Proceedings, Part V 14}, pages 382--398. Springer, 2016.

\bibitem{anderson2018bottom}
Peter Anderson, Xiaodong He, Chris Buehler, Damien Teney, Mark Johnson, Stephen Gould, and Lei Zhang.
\newblock Bottom-up and top-down attention for image captioning and visual question answering.
\newblock In {\em Proceedings of the IEEE conference on computer vision and pattern recognition}, pages 6077--6086, 2018.

\bibitem{huang2019attention}
Lun Huang, Wenmin Wang, Jie Chen, and Xiao-Yong Wei.
\newblock Attention on attention for image captioning.
\newblock In {\em Proceedings of the IEEE/CVF international conference on computer vision}, pages 4634--4643, 2019.

\bibitem{radford2021learningtransferablevisualmodels}
Alec Radford, Jong~Wook Kim, Chris Hallacy, Aditya Ramesh, Gabriel Goh, Sandhini Agarwal, Girish Sastry, Amanda Askell, Pamela Mishkin, Jack Clark, Gretchen Krueger, and Ilya Sutskever.
\newblock Learning transferable visual models from natural language supervision, 2021.

\bibitem{agrawal2019nocaps}
Harsh Agrawal, Karan Desai, Yufei Wang, Xinlei Chen, Rishabh Jain, Mark Johnson, Dhruv Batra, Devi Parikh, Stefan Lee, and Peter Anderson.
\newblock nocaps: novel object captioning at scale.
\newblock In {\em Proceedings of the IEEE International Conference on Computer Vision}, pages 8948--8957, 2019.

\bibitem{Rotstein_2024_WACV}
Noam Rotstein, David Bensa{\"\i}d, Shaked Brody, Roy Ganz, and Ron Kimmel.
\newblock Fusecap: Leveraging large language models for enriched fused image captions.
\newblock In {\em Proceedings of the IEEE/CVF Winter Conference on Applications of Computer Vision (WACV)}, pages 5689--5700, January 2024.

\end{thebibliography}
}

\end{document}